\documentclass{article}

\usepackage{PRIMEarxiv}

\usepackage[utf8]{inputenc} 
\usepackage[T1]{fontenc}    
\usepackage{hyperref}       
\usepackage{url}            
\usepackage{booktabs}       
\usepackage{amsfonts}       
\usepackage{nicefrac}       
\usepackage{microtype}      
\usepackage{lipsum}
\usepackage{graphicx}
\usepackage[numbers,sort&compress]{natbib}          
\usepackage{amssymb}
\usepackage{amsmath}
\usepackage{tabto}
\usepackage[english]{babel}
\usepackage{blindtext}
\usepackage[T1]{fontenc}
\usepackage{amsthm}
\usepackage{multirow}
\usepackage{tabu}
\usepackage{booktabs}
\usepackage[utf8]{inputenc} 
\usepackage{amsfonts}       
\usepackage{graphicx}
\usepackage{bm}
\usepackage[switch]{lineno}
\usepackage{comment}
\usepackage{float} 
\usepackage{algorithm,algorithmic}
\usepackage{subcaption} 
\usepackage{textcomp}
\usepackage{soul, color}
\usepackage{url}
\usepackage{multirow}
\usepackage[normalem]{ulem}
\usepackage{xcolor}
\usepackage{pifont}   
\usepackage{tikz}
\usepackage{threeparttable}
\usepackage{rotating}
\usepackage{afterpage}
\usepackage{placeins}
\usepackage{multicol,multirow}

\graphicspath{{media/}}     

\newcommand{\norm}[1]{\left\lVert #1 \right\rVert}
\newcommand{\ip}[2]{\left\langle #1, #2 \right\rangle}
\newcommand{\cmark}{\ding{51}} 
\newcommand{\xmark}{\ding{55}} 
  
\title{Decoupling Wavelet Sub-bands for Single Source Domain Generalization in Fundus Image Segmentation
}

\author{
  Shramana Dey \\
  Indian Statistical Institute \\
  Kolkata, West Bengal, India \\
  \texttt{shramanadey96@gmail.com}
  \And
  Varun Ajith \\
  Indian Statistical Institute \\
  Kolkata, West Bengal, India \\
  \texttt{varunajith29@gmail.com}
  \And
  Abhirup Banerjee \\
  Department of Engineering Science, University of Oxford \\
  Radcliffe Department of Medicine, University of Oxford \\
  Oxford, United Kingdom \\
  \texttt{abhirup.banerjee@eng.ox.ac.uk}
  \And
  Sushmita Mitra \\
  Indian Statistical Institute \\
  Kolkata, West Bengal, India \\
  \texttt{sushmita@isical.ac.in}
}

\begin{document}
\maketitle

\begin{abstract}
Domain generalization in fundus imaging  is challenging due to variations in acquisition conditions across devices and clinical settings. The inability to adapt to these variations causes performance degradation on unseen domains for deep learning models. Besides, obtaining annotated data across domains is often expensive and privacy constraints restricts their availability.  Although single-source domain generalization (SDG) offers a realistic solution to this problem, the existing approaches frequently fail to capture anatomical topology or decouple appearance from anatomical features. This research introduces \textit{WaveSDG}, a new wavelet-guided segmentation network for SDG. It decouples anatomical structure from domain-specific appearance through a wavelet sub-band decomposition. A novel Wavelet-based Invariant Structure Extraction and Refinement (WISER) module is proposed to process encoder features by leveraging distinct semantic roles of each wavelet sub-band. The module refines low-frequency components to anchor global anatomy, while selectively enhancing directional edges and suppressing noise within the high-frequency sub-bands. Extensive ablation studies validate the effectiveness of the WISER module and its decoupling strategy. Our evaluations on optic cup and optic disc segmentation across one source and five unseen target datasets show that WaveSDG consistently outperforms seven state-of-the-art methods. Notably, it achieves the best balanced Dice score and lowest 95th percentile Hausdorff distance with reduced variance, indicating improved accuracy, robustness, and cross-domain stability.
\end{abstract}


\section{Introduction}
\label{sec:introduction}

Fundus imaging is a gold standard modality for detailed non-invasive assessment of the retina.  It plays a critical role in the early detection of multiple vision-threatening diseases, typically revealing changes in the structural characteristics and formations of lesions in key retinal regions like the optic cup (OC), optic disc (OD), macula, and vascular structures \citep{kumar2023fundus}. These need to be monitored to assess progression of diseases such as glaucoma, papilledema, optic neuritis, and diabetic retinopathy. Unlike invasive techniques, it is safe, inexpensive, and widely accessible, and therefore suitable for large-scale ophthalmic screening, particularly in resource-constrained settings. 
Despite its suitability, large-scale screening places a substantial burden on ophthalmologists needing to review large volumes of images.  Diagnostic interpretation varies across experts due to inherent subjective assessment \citep{kumar2023fundus,shi2024asurvey}. 

Deep learning (DL)-based automated analysis \citep{zeng2025funotta,11428283} has emerged as an important solution  for scaling ophthalmic screening. DL models need a huge amount of high-quality annotated data for training; however, the high cost of expert annotation and stringent privacy regulations restrict their availability and centralized data sharing. These constraints necessitate retraining of models independently on data across clinical centers.  This again becomes computationally expensive and time-consuming. In practice, the performance of most DL models degrade when deployed on data across hospitals or imaging devices, mainly due to domain shift \citep{guan2021domain,yoon2024domain}. It arises from variations in acquisition conditions, such as illumination, image size, texture or sensor noise. Such appearance or style variations cause distribution mismatch between training and deployment data, leading to misalignment of features, spurious pathological cues, and/or overconfident incorrect predictions.

Different strategies for mitigating performance degradation, under domain shift between training (source) and unseen testing (target) data, have been explored. These primarily differ on their degree of reliance on target-domain data. Domain Adaptation (DA) \citep{guan2021domain} aligns feature distributions across source and target domains while assuming access to target data during training, thereby, limiting their real-world clinical deployment. Test-Time Adaptation (TTA) \citep{zeng2025funotta} adapts trained models to target domains using unlabeled samples at inference time, at the expense of increased deployment complexity and stability-reproducibility concerns. In contrast, Domain Generalization (DG) \citep{ren2026fedca,10780969,chen2023treasure,yoon2024domain,ren2025integrated,zhang2020generalizing,ouyang2022causality,su2023rethinking,huang2025ada,wang2025teacher,wang2025hybrid,hu2023devil,jiang2025structure,chen2022maxstyle,xu2022adversarial,safdari2025mixstyleflow,gu2023train,kervadec2019boundary,li2023frequency,qiao2024medical,zhao2024morestyle} seeks to learn representations that generalize to unseen target domains without requiring any access. These are broadly categorized into Multi-source Domain Generalization (MDG) \citep{chen2023treasure,ren2025integrated} and Single-source Domain Generalization (SDG) \citep{zhang2020generalizing,ouyang2022causality,su2023rethinking,huang2025ada,wang2025teacher,wang2025hybrid,hu2023devil,jiang2025structure,chen2022maxstyle,xu2022adversarial,safdari2025mixstyleflow,gu2023train,kervadec2019boundary,li2023frequency,qiao2024medical,zhao2024morestyle}. MDG uses labeled data from multiple source domains to learn domain-invariant representations, but it remains impractical in medical imaging due to annotation costs, data-sharing limitations, and regulatory constraints. SDG represents the most realistic yet challenging scenario, in which models trained on a single source domain must generalize to unseen targets. 

Existing SDG methods attempt to increase style diversity in the source domain during training. Data augmentation approaches \citep{zhang2020generalizing,ouyang2022causality,su2023rethinking,huang2025ada} apply transformations to synthetically replicate style variations. While BigAug \citep{zhang2020generalizing} and Causality inspired Single source Domain Generalization (CSDG) \citep{ouyang2022causality} employ a fixed set of transformations across samples, Saliency-balancing Location-scale Augmentation (SLAug) \citep{su2023rethinking} and Adaptive Augmentation Framework (ADA) \citep{huang2025ada} learn sample-specific augmentation parameters. Nevertheless, most augmentation strategies fail to disentangle domain-specific appearance variations from clinically relevant anatomical structures on medical images. These transformations can inadvertently corrupt structural information unless explicitly constrained \citep{wang2025teacher}. 

Representation learning approaches attempt to disentangle style and content representations using various objectives. Contrastive learning or disentanglement strategies, such as Dual-augmentation Constraint Framework (DCON) \citep{wang2025hybrid}, Channel-level Contrastive Single Domain Generalization (C$^2$SDG) \citep{hu2023devil}, and Pixel-level Constrastive Single Source Domain Generalization (PCSDG) \citep{jiang2025structure}, aim to isolate style-related features during training. Note that, reliably disentangling anatomical structure from domain specific appearance remains challenging, as these factors are inherently coupled in medical images. Another line of research employs adversarial strategies to induce appearance variations during training, thereby, increasing robustness to style shifts. MaxStyle \citep{chen2022maxstyle} perturbs feature distributions to improve robustness against style variations. Adversarial Domain Synthesizer (ADS) \citep{xu2022adversarial} introduces a texture synthesizer guided by a style module, and enforces semantic consistency through mutual information regularization. MoreStyle \citep{zhao2024morestyle} generates realistic and diverse style variation using adversarial training. MixStyleFlow \citep{safdari2025mixstyleflow} further extends ADS through flow-based style mixing; however, they incur high computational cost and optimization instability, limiting their practicality.
Although edge-guided strategies like Edge-Guided Single-source Domain Generalization (EGSDG) \citep{gu2023train} incorporate classical edge detectors to enhance structural awareness, they are unreliable under cross-site imaging variations. This highlights the need for feature-driven edge representation that emphasizes anatomically meaningful edges while suppressing unreliable noisy edges.

Spectral representation provides an alternative perspective to the spatial domain for addressing domain shift. Fourier domain strategies \citep{hu2023devil,li2023frequency,qiao2024medical,zhao2024morestyle} exploit the observation that low-frequency amplitude captures global appearance statistics, while phase preserves semantic structure. For example,  C$^2$SDG \citep{hu2023devil} leverages this property directly for augmentation by swapping low-frequency amplitudes. Frequency-mixed Single-source Domain Generalization (FreeSDG) \citep{li2023frequency}, Random Amplitude Spectrum Synthesis for Single-Source Domain Generalization (RAS$^4$DG) \citep{qiao2024medical}, and MoreStyle \citep{zhao2024morestyle} manipulate frequency spectra to reduce inter-domain shift. Yet, Fourier representation lacks spatial locality, which limits its ability to model spatial organization of anatomical structures. 

\begin{figure}
    \centering
    \includegraphics[width=0.6\columnwidth]{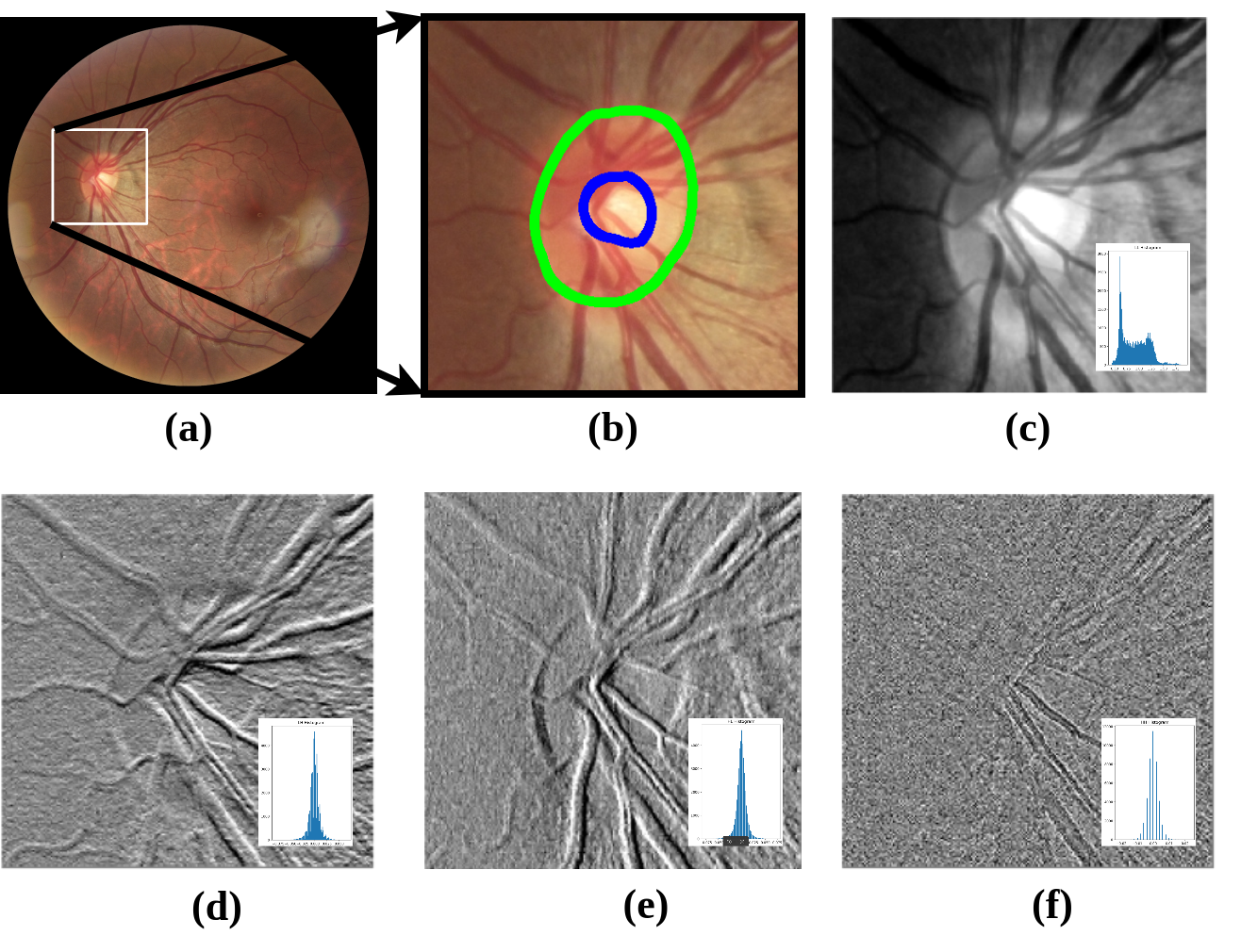}
    \caption{Representative wavelet decomposition of (a) Fundus image, with (b) extracted \textcolor{green}{Optic Disc}-\textcolor{blue}{Optic Cup} regions, and its corresponding representation in wavelet sub-bands (c) LL, (d) LH, (e) HL, and (f) HH.}
    \label{fig:dwtdecom}
\end{figure}

The Discrete Wavelet Transform (DWT) \citep{finder2024wavelet} offers a principled alternative to jointly analyze spatial structure and frequency characteristics of image features. Wavelet decomposition separates features into a low-frequency sub-band (LL) to capture global structural context and high-frequency sub-bands (LH, HL, HH) for encoding directional edge responses \citep{zhang2025uwt,deng2019wavelet}. This decomposition naturally aligns with anatomical structure and imaging properties in retinal fundus images, with LL predominantly preserving anatomical layout \citep{zhang2025uwt}. LH, HL capture horizontal and vertical edges, and HH sub-band is often dominated by sensor noise and acquisition artifacts \citep{deng2019wavelet}. 
Figure~\ref{fig:dwtdecom} illustrates a representative sample of the Optic Disc and Optic Cup region. The LL sub-band preserves smooth intensity and clear anatomical structure, as reflected in its broad intensity distribution. On the other hand, the LH and HL sub-bands distinctly highlight the anatomical boundary and vessel. The HH sub-band appears texture-like, with minimal structural information and a sharply peaked near-zero distribution.

Existing wavelet-based segmentation frameworks primarily exploit multi-scale wavelet representation, without leveraging the principled semantic differences among the sub-bands. For example, Wang et al. \citep{wang2024learning} treats all wavelet components as equally informative, thus propagating the noisy HH sub-band during feature learning. 

Motivated by the limitations in existing literature, we propose a novel Wavelet-based Invariant Structure Extraction and Refinement (\textbf{WISER}) module that decomposes encoder features into wavelet sub-bands and refines them based on their structural relevance. The $U$-shaped \citep{ronneberger2015u} segmentation backbone is extended to incorporate the WISER module for single source domain generalization (\textbf{WaveSDG}). This decouples anatomical structures from domain-specific appearance, prior to fusion with the decoder features. 
The contribution of the research is summarized below. 
\begin{itemize}
    \item A novel lightweight, resource-efficient WISER module is proposed. It replaces the identity skip connections in the $U$-shaped segmentation backbone for effective decoupling of anatomical structure from appearance cues to produce structure-aware features under domain shift.
    \begin{itemize}
    \item We introduce wavelet sub-bands to disentangle anatomy from style, decomposing encoder features and refining based on structural relevance. 
    \item We introduce complementary Edge Booster (EB) and Edge Selector (ES)  to selectively refine and localize structural boundaries. This facilitates precise delineation of Region of Interest (ROI), even under large domain-shift.
    \end{itemize} 
    \item A new segmentation architecture WaveSDG, incorporating WISER module, is designed for SDG, to prevent  propagation of domain-specific appearance cues through skip connections.
    \item Extensive experiments across one source dataset (Refuge \citep{refuge2020paper}) and five unseen target datasets (Drishti-GS \citep{drishti_gs}, Gamma \citep{wu2023gamma}, and Chákṣu \citep{chaksu} encompassing Bosch, Forus, and Remidio) demonstrate that WaveSDG consistently achieves the lowest HD95, indicating high boundary precision and stability in most target datasets.
    \item The WISER module yields the highest balanced Dice coefficient for OD/OC segmentation across both clinical-grade as well as handheld fundus cameras, out-performing the state-of-the-art (SOTA) models. 
\end{itemize}
The rest of the manuscript is organized as follows. Section~\ref{sec:method} introduces the WaveSDG architecture, including the WISER module, loss function, training strategy, datasets used, and experimental details. Section~\ref{sec:exp&res} summarizes the comprehensive qualitative and quantitative evaluation, encompassing ablation and interpretability of each component involved in WISER module. Finally, Section~\ref{sec:conclusion} concludes the manuscript.

\section{Methodology}
\label{sec:method}

\begin{figure*}
    \centering
    \includegraphics[width=1.0\textwidth]{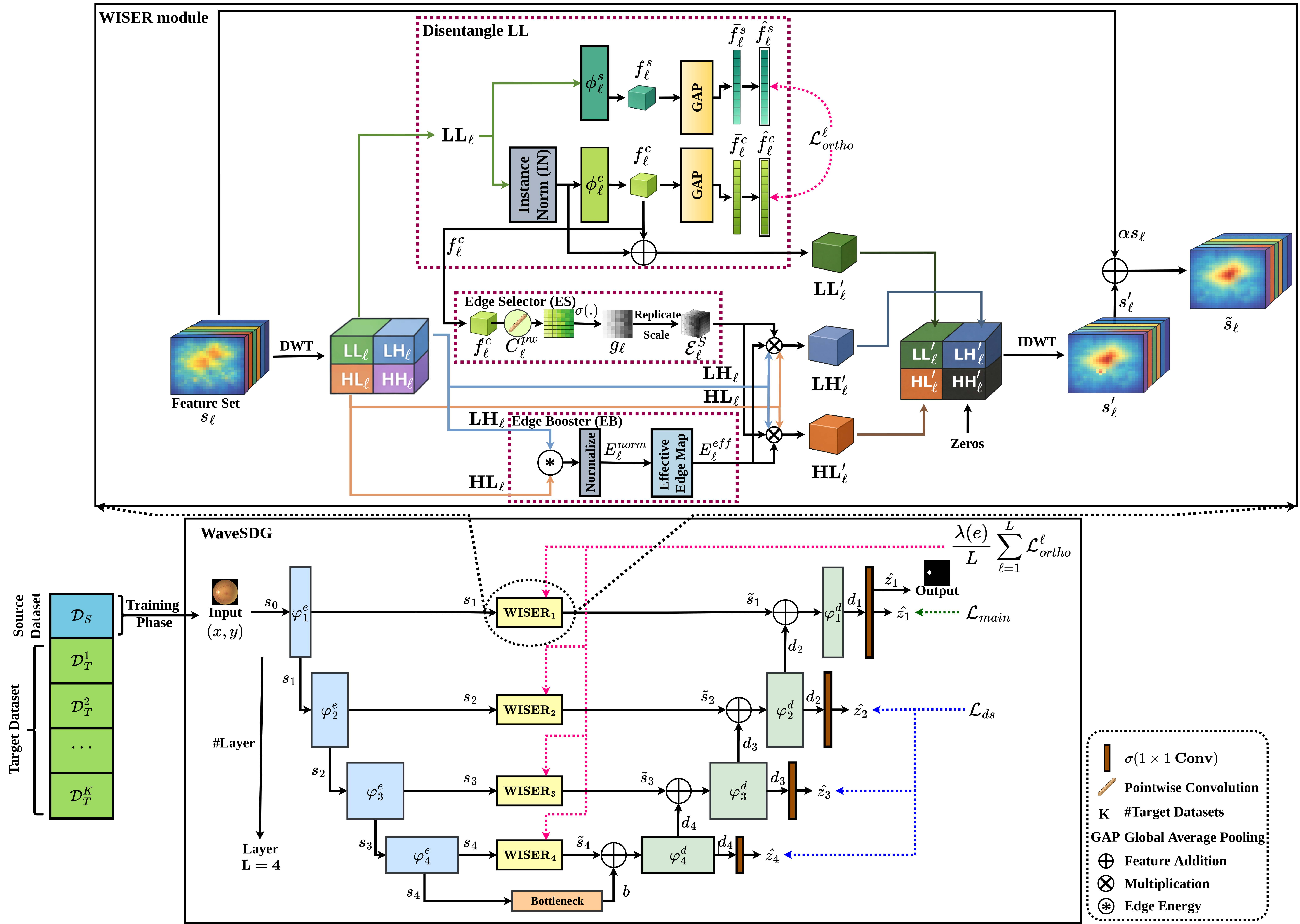}
    \caption{The proposed WaveSDG architecture. The lower and upper rectangles illustrate the overall network and the novel WISER module, respectively. The model is trained exclusively on the source dataset, while the target dataset is used only during evaluation. The WISER module filters the encoder features prior to it being passed to the decoder. Dotted rectangles highlight the internal components of the WISER module. Solid lines denote forward propagation path and the dotted lines represent the point from which gradients flow during back-propagation starts.}
    \label{fig:wiserseg}
\end{figure*}

The proposed \textbf{W}avelet-based \textbf{I}nvariant \textbf{S}tructure \textbf{E}xtraction and \textbf{R}efinement (WISER) module incorporated in a segmentation backbone (WaveSDG) of Fig.~\ref{fig:wiserseg} considers a source domain $\mathcal{D}_S = \{(x_i, y_i)\}_{i=1}^{N}$, where each input image $x_i \in \mathbb{R}^{3 \times H \times W}$ corresponds to a segmentation mask $y_i \in \{0,1, \cdots, M\}^{H \times W}$ for height $H$, width $W$, and $M$ classes. Here, $N$ denotes the number of samples in the source domain. The goal is to train a segmentation network on $\mathcal{D}_S$ that can generalize its performance to any unseen target domain $\mathcal{D}_T$. The $k$-th target domain is defined as $\mathcal{D}_T^k = \{(x_j^k, y_j^k)\}_{j=1}^{N_T^k}$ containing $N_T^k$ labeled samples. The WaveSDG model integrates a segmentation backbone with a set of anatomy-preserving style-filtering WISER modules.

The segmentation backbone adopts a generic encoder-decoder architecture. The encoder contains $L$ layers and progressively extracts hierarchical representations $s_\ell = \varphi_\ell^e{(s_{\ell-1})}$ from the input $x$ using convolution layer $\varphi$, with $s_0 = x$. Each of these feature maps $s_\ell$ has dimension $\mathbb{R}^{C^\ell \times \frac{H}{2^\ell} \times \frac{W}{2^\ell}}$. While the shallow layers capture fine spatial details, the deeper layers encode abstract but semantically meaningful features. 

The WISER module for layer $\ell$ processes the output of each encoder layer to generate filtered features $\tilde{s}_\ell$. The final encoder representation $s_L$ passes through a bottleneck to produce $b$. The decoders progressively restore spatial resolution by up-sampling between consecutive layers. The decoder output is computed as $d_\ell = \varphi_\ell^d{(d_{\ell+1},\tilde{s}_{\ell})}$, with  $d_L = \varphi_L^d{(b,\tilde{s}_L)}$. 

\subsection{WISER module}
\label{subsec:wiser}

Given the feature map $s_\ell$ from encoder layer $\ell$ with dimension $\mathbb{R}^{B \times C^\ell \times \frac{H}{2^\ell} \times \frac{W}{2^\ell}}$, where $B$ denotes the batch size, the simple and computationally-efficient depthwise Haar DWT decomposes $s_\ell$ into four sub-bands
\begin{equation}
\label{eq:decompose_dwt}
(LL_\ell,LH_\ell,HL_\ell,HH_\ell)=\mathrm{DWT}(s_\ell),
\end{equation}
with $LL_\ell,LH_\ell,HL_\ell,HH_\ell \in \mathbb{R}^{B \times C^\ell \times \frac{H}{2^{\ell+1}} \times \frac{W}{2^{\ell+1}}}$.
A depthwise Inverse-DWT (IDWT) reconstructs the corresponding filtered feature $s_\ell'$, from the modified sub-bands, as
\begin{equation}
\label{eq:idwt}
s_\ell' = IDWT(LL_\ell^{'},LH_\ell^{'},HL_\ell^{'},HH_\ell^{'}).
\end{equation}

\subsubsection{Learning anatomy from LL sub-band}
\label{subsubsec:ll_processing}

As the LL sub-band captures low-frequency components of the feature map, it encodes global anatomical structures important for ROI localization. It also encompasses domain-specific appearance cues, such as illumination, color tone, and shading \citep{deng2019wavelet,zhang2025uwt}. $LL_\ell$ is decomposed into a pair of decorrelated  feature representations, involving style and anatomy (content), as
\begin{equation}
\label{eq:style_content}
f_\ell^s=\phi_\ell^s(LL_\ell), \qquad f_\ell^c=\phi_\ell^c(\mathrm{IN}(LL_\ell)).
\end{equation}
Here $\phi_\ell^s(\cdot)$ and $\phi_\ell^c(\cdot)$ correspond to identical convolution layers having two stacked $3\times 3$ ConvReLU blocks, and $\mathrm{IN}$ denotes Instance Normalization. The $\mathrm{IN}$ reduces sensitivity to per-image intensity statistics and stabilizes content extraction.
The Global Average Pooling (GAP) individually flattens $f_\ell^s$ and $f_\ell^c$ feature maps to produce  $\bar{f}_\ell^s$ and $\bar{f}_\ell^c$, respectively. 
The training objective optimizes an orthogonal loss $\mathcal{L}_{ortho}^\ell$ between $f_\ell^s$ and $f_\ell^c$ to enforce feature decorrelation and reduced shared information between them. It is computed as
\begin{equation}
\mathcal{L}_{ortho}^\ell=\frac{1}{B}\sum_{i=1}^{B}\left[\left(\ip{\hat{f}_\ell^c}{\hat{f}_\ell^s}\right)^2\right],
\end{equation}
where
$$\hat{f}_\ell^c=\frac{\bar{f}_\ell^c-\mu(\bar{f}_\ell^c)}{\norm{\bar{f}_\ell^c-\mu(\bar{f}_\ell^c)}_2+\epsilon},\quad
\hat{f}_\ell^s=\frac{\bar{f}_\ell^s-\mu(\bar{f}_\ell^s)}{\norm{\bar{f}_\ell^s-\mu(\bar{f}_\ell^s)}_2+\epsilon},$$ 
with $\mu$ representing the mean and $\epsilon$ being a very small number.
The filtered $LL_\ell'$ sub-band is obtained as
\begin{equation}
\label{eq:LL_recon}
    LL_\ell' = \mathrm{IN}(LL_\ell) + f_\ell^c.
\end{equation}

\subsubsection{Selective refinement of LH and HL}
\label{subsubsec:lh_hl_processing}

The LH and HL sub-bands encode the high-frequency horizontal and vertical edge responses in the input. Although the LL sub-band represents the global anatomical structure, it produces blurred boundaries. Thus, the LH and HL sub-bands are exploited to generate precise localization of the ROI boundary. Variation in imaging conditions often degrades image quality, with weakened edge response. The Edge Booster (EB) serves to adaptively amplify edge responses under low contrast.  An Edge Selector (ES) is embedded inside the WISER module to selectively retain edge responses from the LH and HL bands.

\paragraph{Edge Booster} Poor image quality often changes edge response in LH and HL sub-bands globally, causing boundaries to disappear even when the selector identifies the correct location. The booster computes an edge energy, normalized between 0 and 1, as
\begin{equation}
\label{eq:enorm}
E_\ell^{norm}=\frac{1}{C}\sum_{c=1}^{C}\big(|LH_\ell|+|HL_\ell|\big)\in\mathbb{R}^{B\times 1\times \tfrac{H}{2^{\ell+1}}\times \tfrac{W}{2^{\ell+1}}}.
\end{equation}
The EB performs an adaptive thresholding to suppress weak or background edges using $$\tau = \mu(E_\ell^{norm}) + \beta \sigma(E_\ell^{norm}),$$ where $\mu$ is the mean and $\sigma$ the standard deviation. Thus, the effective edge map $E_\ell^{eff}$, obtained after thresholding $E_\ell^{norm}$, preserves the salient high-energy edge structure while removing weak-edge responses (noise).

\paragraph{Edge Selector} LH and HL sub-bands encode both anatomical boundaries and background edges. The selector considers edges aligned with high activations in $f_\ell^c$ as informative. It treats misaligned edges, in $LH_l$ and $HL_l$, as undesired background responses driven by appearance variations. 

A selector predicts a spatial gate using the content feature $f_\ell^c$. The anatomy-aware feature map $f_\ell^c$ is passed through a point-wise convolution $C_\ell^{pw}$, followed by a sigmoid, to estimate spatial regions having high activation corresponding to anatomical structures. We have
\begin{equation}
\label{eq:gl}
g_\ell = \sigma{(C_\ell^{pw}(f_\ell^c))}\in [0,1]^{B\times 1\times \tfrac{H}{2^{\ell+1}}\times \tfrac{W}{2^{\ell+1}}}.
\end{equation}
The resultant spatial gate $g_\ell$ is replicated across the channel dimension, to make it compatible for element-wise multiplication with both $LH_\ell$ and $HL_\ell$. It is converted to centerize at 1 and prevent hard masking as follows:
\begin{equation}
\label{eq:scalefactor}
\mathcal{E}_\ell^\mathcal{S} = 1 + \varepsilon (2g_\ell-1),
\end{equation}
where $\varepsilon$ controls maximum deviation from identity. Therefore, ES suppresses background edges while enhancing informative ones, thereby avoiding complete edge elimination.

The edge selector and edge booster factors simultaneously refine and enhance the edges in $LH_\ell$ and $HL_\ell$ as
\begin{equation}
\label{eq:LH_HL_recon}
LH_\ell' = LH_\ell \odot \mathcal{E}_\ell^\mathcal{S} \odot E_\ell^{eff},\qquad HL_\ell' = HL_\ell \odot \mathcal{E}_\ell^\mathcal{S} \odot E_\ell^{eff},
\end{equation}
where $\odot$ is the element-wise multiplication operator. While the selector targets the edges where evidence should be strengthened, the booster targets how much of the evidence needs to be strengthened there.

\subsubsection{Reconstruction using filtered sub-bands}
\label{subsubsec:recon}

Equations~(\ref{eq:LL_recon}) and (\ref{eq:LH_HL_recon}) define the filtered LL, LH, and HL sub-bands at layer $\ell$. Although the HH sub-band contains high-frequency diagonal edge responses, various domain-specific noises also dominate. Here, the HH sub-band is completely suppressed by setting $HH_\ell' = 0$, with the filtered feature map $s_\ell'$ being constructed using eqn. (\ref{eq:idwt}). 

A residual connection is used to construct $\tilde{s}_\ell$ to improve stability during training. Thus,
\begin{equation}
\label{eq:wavelet_res}
    \tilde{s}_\ell = \alpha s_\ell + s_\ell',
\end{equation}
where $\alpha$ is kept small to prevent over reliance on the unfiltered feature map $s_\ell$ during the training.

\subsection{Training}
\label{subsec:training}

The output of each decoder layer $d_\ell$ is passed through a $1 \times 1$ convolution to produce logit $z_\ell$. Each $z_\ell$ is next passed through a sigmoid activation function to generate $\hat{z}_\ell$. The model uses multi-class Dice loss $\mathcal{L}_{Dice} (\cdot,\cdot)$ \citep{sudre2017generalised} for the main segmentation task. The main loss $\mathcal{L}_{main}$ is calculated as
\begin{equation}
    \mathcal{L}_{main} = \mathcal{L}_{Dice}(\hat{z}_1, y).
\end{equation}
A Deep Supervision (DS) loss $\mathcal{L}_{ds}$ is used to stabilize optimization and encourage coarse to fine boundary consistency. It is computed as
\begin{equation}
    \mathcal{L}_{ds} = \sum_{\ell=2}^{L}w_\ell\mathcal{L}_{Dice}(\text{UP}_\ell(\hat{z}_\ell), y),
\end{equation}
where hyperparameter $w_\ell$ corresponds to the weight of $\mathcal{L}_{Dice}$  at layer $\ell$ and $\text{UP}_\ell$ denotes up-sampling of $\hat{z}_\ell$ by factor $2^\ell$. 
The total loss $\mathcal{L}$ becomes
\begin{equation}
    \mathcal{L} = \mathcal{L}_{main} + \mathcal{L}_{ds} + \frac{\lambda(e)}{L}\sum_{\ell=1}^{L}\mathcal{L}_{ortho}^\ell,
\end{equation}
where $\lambda(e)$ is a linear ramp schedule, with a warmup epoch $e_w$ and a ramp epoch of $e_r$. It is computed as
\begin{equation}
\label{eq:orthoscheduler}
\lambda(e)=
\begin{cases}
0, & e\le e_w,\\
\lambda_{\max}\cdot \dfrac{e-e_w}{e_r}, & e_w < e < e_w+e_r,\\
\lambda_{\max}, & e\ge e_w+e_r.
\end{cases}
\end{equation}
This scheduler enables the model to first learn the strong, task-specific representations, followed by a progressive incorporation of the orthogonality constraint. This results in a stable and effective feature decoupling.

\subsection{Datasets}
\label{subsec:dataset}
The proposed WaveSDG model was trained on a single publicly available fundus dataset Refuge \citep{refuge2020paper} ($N = 1200$ images), the souce domain, which provides joint OD/OC pixel-wise annotation. The evaluation protocol used five publicly available datasets as target domains, to assess the SDG for OD/OC segmentation. These include Drishti-GS \citep{drishti_gs} ($N_T^1 = 101$ images), Gamma \citep{wu2023gamma} ($N_T^2 =100$ images), and the multi-domain Chákṣu \citep{chaksu} datasets. The Chákṣu consists of three distinct acquisition settings, {\it viz.} Bosch ($N_T^3 =1074$), Forus ($N_T^4 =145$), and Remidio ($N_T^5 =126$). Table~\ref{tab:dataset} summarizes the characteristics of the datasets used.

\begin{table}
\centering
\caption{Source and Target Domain datasets.}
\label{tab:dataset}
\renewcommand{\arraystretch}{1.25}
\resizebox{0.6\columnwidth}{!}{%
\begin{tabular}{|lll|c|c|c|l|}
\hline
\multicolumn{3}{|l|}{\textbf{Dataset}}                                                                                                                   & \textbf{\begin{tabular}[c]{@{}c@{}}Size \\ (pixels)\end{tabular}}                & \textbf{\begin{tabular}[c]{@{}c@{}}\# Samples\\ (Train-Val-Test)\end{tabular}} & \textbf{\begin{tabular}[c]{@{}c@{}}Field-of-\\ View \end{tabular}} & \textbf{Camera}                                                         \\ \hline
\multicolumn{3}{|l|}{\begin{tabular}[c]{@{}l@{}}\textbf{Source} --\\ Refuge\end{tabular}}                                                             & \begin{tabular}[c]{@{}c@{}}2124 $\times$ 2056,\\ 1634 $\times$ 1634\end{tabular} & 840-120-240                                                                    & 40$^\circ$                                                                                               & \begin{tabular}[c]{@{}l@{}}Zeiss Visucam 500,\\ Canon CR-2\end{tabular} \\ \hline
\multicolumn{1}{|l|}{\multirow{5}{*}{\rotatebox{90}{\textbf{Target}}}} & \multicolumn{2}{l|}{Drishti-GS}                                                 & 2896 $\times$ 1944                                                               & 0-0-101                                                                        & 30$^\circ$                                                                                               & -                                                                       \\ \cline{2-7} 
\multicolumn{1}{|l|}{}                                                 & \multicolumn{2}{l|}{Gamma}                                                      & \begin{tabular}[c]{@{}c@{}}2000 $\times$ 2992,\\ 1934 $\times$ 1956\end{tabular} & 0-0-100                                                                        & -                                                                                                & \begin{tabular}[c]{@{}l@{}}KOWA,\\ Topcon TRC-NW400\end{tabular}        \\ \cline{2-7} 
\multicolumn{1}{|l|}{}                                                 & \multicolumn{1}{l|}{\multirow{3}{*}{\rotatebox{90}{\textbf{Chákṣu}}}} & Bosch   & 1920 $\times$ 1440                                                               & 0-0-145                                                                        & 40$^\circ$                                                                                               & Handheld Fundus Camera                                                  \\ \cline{3-7} 
\multicolumn{1}{|l|}{}                                                 & \multicolumn{1}{l|}{}                                                 & Forus   & 2048 $\times$ 1536                                                               & 0-0-126                                                                        & 40$^\circ$                                                                                               & 3Nethra Classic                                                         \\ \cline{3-7} 
\multicolumn{1}{|l|}{}                                                 & \multicolumn{1}{l|}{}                                                 & Remidio & 2448 $\times$ 3264                                                               & 0-0-1074                                                                       & 40$^\circ$                                                                                               & Fundus-on-Phone                                                         \\ \hline
\end{tabular}%
}
\end{table}


\begin{table*}
\centering
\caption{Comparative performance of WaveSDG and seven SOTA models, in joint OD/OC segmentation, using Refuge as the source dataset. Metrics reported as mean ± standard deviation, with \textbf{bold} representing best score and \ul{underline} second-best score.}
\label{tab:quantitativeRes}
\resizebox{\textwidth}{!}{
\renewcommand{\arraystretch}{1.5}
\begin{tabular}{|lll|l|l|l|l|l|l|l|l|l|}
\hline
\multicolumn{3}{|l|}{\textbf{Dataset}}                                                                                                                                      & \textbf{Metric}      & \textbf{WaveSDG}       & \textbf{BigAug}       & \textbf{SLAug}  & \textbf{$\mathbf{C^2SDG}$} & \textbf{PCSDG}     & \textbf{FreeSDG} & \textbf{MoreStyle}     & \textbf{EGSDG}  \\ \hline
\multicolumn{3}{|l|}{\multirow{4}{*}{\begin{tabular}[c]{@{}l@{}}\textbf{Source} -- \\ Refuge\end{tabular}}}                                                                 & \textbf{$DSC_{OD}$}   & \textbf{91.21 ± 4.23}\(^{*}\)          & \ul{90.77 ± 5.77}       & 85.41 ± 8.05 & 89.53 ± 5.12   & 86.23 ± 8.78    & 85.09 ± 11.36  & {85.01± 8.92}   & 86.83 ± 7.35 \\ \cline{4-12}
\multicolumn{3}{|l|}{}                                                                                                                                                      & \textbf{$DSC_{OC}$}   & \textbf{89.76 ± 5.57}\(^{*}\) & 87.31 ± 6.19       & 84.33 ± 7.92 & {\ul{88.20 ± 6.71}}      & 83.75 ± 10.44    & 84.45 ± 8.19  & 85.98 ± 7.12        & 85.70 ± 7.50 \\ \cline{4-12}
\multicolumn{3}{|l|}{}                                                                                                                                                      & \textbf{$HD95_{OD}$} & \textbf{3.86 ± 2.20}\(^{*}\)     & 4.98 ± 4.93           & 6.54 ± 3.38     & {\ul{4.02 ± 2.23}}          & 6.70 ± 13.91       & 7.34 ± 13.47     & 8.03 ± 3.45            & 5.69 ± 3.10     \\ \cline{4-12}
\multicolumn{3}{|l|}{}                                                                                                                                                      & \textbf{$HD95_{OC}$} & \textbf{3.68 ± 1.78}\(^{*}\)     & 5.26 ± 4.47           & 5.49 ± 2.86     & {\ul{4.31 ± 2.38}}          & 5.84 ± 2.97        & 12.11 ± 10.84    & 5.13 ± 2.34            & 5.16 ± 2.76     \\ \hline
\multicolumn{1}{|l|}{\multirow{20}{*}{\rotatebox{90}{\textbf{Target}}}} & \multicolumn{2}{l|}{\multirow{4}{*}{DrishtiGS}}                                                   & \textbf{$DSC_{OD}$}   & \textbf{80.57 ± 9.30}          & 66.87 ± 17.09       & 76.51 ± 14.71 & \ul{80.45 ± 10.54}   & 66.95 ± 11.29    & 35.89 ± 15.47  & {74.20 ± 11.74}  & 42.16 ± 24.80 \\ \cline{4-12}
\multicolumn{1}{|l|}{}                                                  & \multicolumn{2}{l|}{}                                                                             & \textbf{$DSC_{OC}$}   & \textbf{82.06 ± 11.59} & 63.51 ± 25.08       & 60.51 ± 18.82 & {\ul{80.73 ± 14.33}}      & 62.97 ± 25.30    & 32.57 ± 16.79  & 72.78 ± 22.56        & 45.90 ± 24.67 \\ \cline{4-12}
\multicolumn{1}{|l|}{}                                                  & \multicolumn{2}{l|}{}                                                                             & \textbf{$HD95_{OD}$} & \textbf{9.66 ± 3.97}     & 55.92 ± 39.03         & 47.29 ± 79.63   & {\ul{9.73 ± 5.91}}          & 37.70 ± 65.43      & 63.48 ± 59.28    & 10.33 ± 9.64           & 59.32 ± 66.65   \\ \cline{4-12}
\multicolumn{1}{|l|}{}                                                  & \multicolumn{2}{l|}{}                                                                             & \textbf{$HD95_{OC}$} & \textbf{11.44 ± 5.83}    & 34.36 ± 43.37         & 27.43 ± 38.87   & 16.79 ± 12.01              & 18.20 ± 7.75       & 62.61 ± 74.04    & {\ul{15.62 ± 8.72}}     & 51.60 ± 61.18   \\ \cline{2-12}
\multicolumn{1}{|l|}{}                                                  & \multicolumn{2}{l|}{\multirow{4}{*}{Gamma}}                                                       & \textbf{$DSC_{OD}$}   & {\textbf{81.49 ± 10.59}}    & \ul{75.66 ± 16.75}       & 47.18 ± 27.57 & 60.60 ± 13.12   & 52.84 ± 16.49    & 48.94 ± 25.41  & 72.53 ± 16.73        & 56.98 ± 18.68 \\ \cline{4-12}
\multicolumn{1}{|l|}{}                                                  & \multicolumn{2}{l|}{}                                                                             & \textbf{$DSC_{OC}$}   & \textbf{79.52 ± 11.02}\(^{*}\) & {\ul{68.16 ± 22.34}} & 52.81 ± 22.99 & 65.85 ± 18.34            & 50.75 ± 29.85    & 38.39 ± 16.02  & 66.23 ± 18.39        & 54.15 ± 26.81 \\ \cline{4-12}
\multicolumn{1}{|l|}{}                                                  & \multicolumn{2}{l|}{}                                                                             & \textbf{$HD95_{OD}$} & \textbf{10.35 ± 9.33}\(^{*}\)            & 33.37 ± 45.67         & 57.65 ± 85.15   & \ul{20.30 ± 34.13}     & 32.38 ± 42.12      & 28.47 ± 31.25    & {24.85 ± 43.63}    & 25.33 ± 24.70   \\ \cline{4-12}
\multicolumn{1}{|l|}{}                                                  & \multicolumn{2}{l|}{}                                                                             & \textbf{$HD95_{OC}$} & {\textbf{7.42 ± 7.01}}\(^{*}\)      & 32.50 ± 56.09         & 25.86 ± 47.26   & {20.43 ± 35.13}        & 31.88 ± 24.35      & 28.62 ± 29.97    & \ul{17.47 ± 25.37} & 27.71 ± 28.84   \\ \cline{2-12}
\multicolumn{1}{|l|}{}                                                  & \multicolumn{1}{l|}{\multirow{12}{*}{\rotatebox{90}{\textbf{Chákṣu}}}} & \multirow{4}{*}{Bosch}   & \textbf{$DSC_{OD}$}   & \textbf{90.07 ± 2.95}          & \ul{89.98 ± 1.04}       & 66.81 ± 19.27 & 89.22 ± 4.43   & 83.61 ± 5.92    & 42.45 ± 21.75  & {75.68 ± 7.63}  & 78.37 ± 10.57 \\ \cline{4-12}
\multicolumn{1}{|l|}{}                                                  & \multicolumn{1}{l|}{}                                                  &                          & \textbf{$DSC_{OC}$}  & \textbf{84.68 ± 7.52} & 67.78 ± 3.93       & 76.72 ± 11.89 & {\ul{83.08 ± 10.37}}      & 59.54 ± 29.41    & 31.06 ± 10.34  & 68.25 ± 22.98        & 66.47 ± 19.29 \\ \cline{4-12}
\multicolumn{1}{|l|}{}                                                  & \multicolumn{1}{l|}{}                                                  &                          & \textbf{$HD95_{OD}$} & \textbf{3.98 ± 1.40}     & 7.29 ± 8.34           & 88.72 ± 81.62   & {\ul{7.21 ± 6.40}}          & 12.59 ± 24.71      & 42.24 ± 32.73    & 9.16 ± 26.34           & 10.26 ± 11.51   \\ \cline{4-12}
\multicolumn{1}{|l|}{}                                                  & \multicolumn{1}{l|}{}                                                  &                          & \textbf{$HD95_{OC}$} & \textbf{4.96 ± 1.70}     & 7.87 ± 5.57           & 45.36 ± 69.99   & 14.9 ± 26.75               & 8.54 ± 3.20        & 36.80 ± 49.61    & {\ul{6.93 ± 2.38}}      & 9.17 ± 4.47     \\ \cline{3-12}
\multicolumn{1}{|l|}{}                                                  & \multicolumn{1}{l|}{}                                                  & \multirow{4}{*}{Forus}   & \textbf{$DSC_{OD}$}   & \textbf{90.63 ± 2.49}\(^{*}\)          & 81.75 ± 3.18       & 72.66 ± 22.31 & \ul{84.97 ± 2.34}   & 82.96 ± 6.92    & 53.33 ± 22.24  & {67.61± 15.03}   & 69.76 ± 16.01 \\ \cline{4-12}
\multicolumn{1}{|l|}{}                                                  & \multicolumn{1}{l|}{}                                                  &                          & \textbf{$DSC_{OC}$}   & \textbf{87.99 ± 5.89} & 69.23 ± 3.79       & 72.89 ± 19.35 & {\ul{83.60 ± 7.47}}      & 64.91 ± 27.06    & 41.24 ± 5.31  & 74.15 ± 18.71          & 66.06 ± 22.51 \\ \cline{4-12}
\multicolumn{1}{|l|}{}                                                  & \multicolumn{1}{l|}{}                                                  &                          & \textbf{$HD95_{OD}$} & \textbf{5.14 ± 1.42}\(^{*}\)     & 14.76 ± 7.94         & 81.33 ± 116.41  & 21.27 ± 48.71              & {\ul{10.85 ± 4.34}} & 52.90 ± 45.75    & 12.80 ± 20.15          & 18.10 ± 18.02   \\ \cline{4-12}
\multicolumn{1}{|l|}{}                                                  & \multicolumn{1}{l|}{}                                                  &                          & \textbf{$HD95_{OC}$} & \textbf{5.41 ± 1.96}\(^{*}\)     & 28.74 ± 36.60         & 29.87 ± 62.69   & 13.46 ± 25.95              & 12.18 ± 4.98       & 52.11 ± 62.02    & {\ul{10.86 ± 3.73}}     & 13.25 ± 7.55    \\ \cline{3-12}
\multicolumn{1}{|l|}{}                                                  & \multicolumn{1}{l|}{}                                                  & \multirow{4}{*}{Remidio} & \textbf{$DSC_{OD}$}   & \ul{85.29 ± 12.72}          & 74.26 ± 9.83       & 52.53 ± 34.29 & \textbf{87.05 ± 3.19}   & 71.12 ± 18.20    & 56.97 ± 11.26  & {70.54 ± 18.09}  & 72.64 ± 18.25 \\ \cline{4-12}
\multicolumn{1}{|l|}{}                                                  & \multicolumn{1}{l|}{}                                                  &                          & \textbf{$DSC_{OC}$}   & \textbf{82.97 ± 12.26}\(^{*}\) & 63.83 ± 10.48       & 53.98 ± 32.77 & {\ul{80.49 ± 10.37}}      & 55.59 ± 29.53    & 46.49 ± 8.12  & 70.60 ± 19.82        & 67.26 ± 21.90 \\ \cline{4-12}
\multicolumn{1}{|l|}{}                                                  & \multicolumn{1}{l|}{}                                                  &                          & \textbf{$HD95_{OD}$} & \textbf{6.35 ± 4.69}\(^{*}\)     & 28.43 ± 26.47         & 148.06 ± 144.78 & {\ul{7.45 ± 24.87}}         & 14.03 ± 13.12      & 43.78 ± 37.61    & 16.04 ± 27.91            & 13.12 ± 15.59   \\ \cline{4-12}
\multicolumn{1}{|l|}{}                                                  & \multicolumn{1}{l|}{}                                                  &                          & \textbf{$HD95_{OC}$} & \textbf{6.74 ± 4.05}\(^{*}\)     & 27.15 ± 21.54         & 122.05 ± 145.21 & {\ul{7.78 ± 4.27}}          & 14.28 ± 6.85       & 42.94 ± 54.02    & 11.75 ± 23.17          & 13.03 ± 8.58    \\ \hline
\end{tabular}}
\vspace{2mm}
{\footnotesize \(^{*}\) indicates statistically significant improvement over the second-best method in terms of one-tailed paired Wilcoxon signed-rank test ($p<0.05$).}
\end{table*}

\subsection{Experimental setup and evaluation metrics}
\label{subsec:expSetup&evalMetric}
The WaveSDG model employed a pretrained ResNet18 \citep{resnet} as the encoder in the segmentation backbone. The pipeline resized each input image to a dimension of $512 \times 512$. The training process used a fixed batch size of 8, and optimized the model for a maximum of 100 epochs using the Adam optimizer with a learning rate of $1 \times 10^{-4}$. It also employed early stopping with patience of 5 epochs, based on the validation performance to obtain the best model. The residual connection in eqn.~(\ref{eq:wavelet_res}) used $\alpha=0.5$. The DS mechanism assigned weights $w_\ell = \{0.2, 0.2, 0.1\}$ for $\ell = \{2, 3, 4\}$, respectively. The scheduling strategy in eqn.~(\ref{eq:orthoscheduler}) regulated the orthogonality loss $\mathcal{L}_{ortho}$ using a warmup phase of 5 epochs, followed by a ramp-up over 10 epochs. The implementation used Python 3.9 with PyTorch, and executed on a single NVIDIA Tesla P6 GPU with 16 GB memory.

Quantitative metrics used in the evaluation were Dice Score Coefficient (DSC) and 95th Percentile Hausdorff Distance (HD95). While $DSC_{class}$ measures the overlap between the ground truth (GT) and the predicted mask, $HD95_{class}$ quantifies the boundary alignment for a particular $class$. A high DSC score indicates better overlap between GT and prediction, whereas a low HD95 signifies a better boundary alignment between GT and the prediction. The Maximum Mean Discrepancy (MMD), Jensen-Shannon Divergence (JSD), and Fréchet Distance \citep{wang2019evolutionary} quantify the generalization capability via kernel mean difference, probabilistic divergence, and Gaussian statistic distance, respectively.

\section{Experimental Results}
\label{sec:exp&res}

This section presents a comparative analysis of our proposed WaveSDG network with related SOTA, along with ablations, computational resource requirement, and qualitative output.

\subsection{Comparison with SOTA}
\label{subsec:com_study}
The proposed WaveSDG was compared with related SOTA, specially designed for medical images. These include augmentation-based (i) BigAug \citep{zhang2020generalizing} and (ii) SLAug \citep{su2023rethinking}, the contrastive feature disentanglement method (iii) C$^2$SDG \citep{hu2023devil}, PCSDG \citep{jiang2025structure}, frequency-based (iv) FreeSDG \citep{li2023frequency}, (v) MoreStyle \citep{su2023rethinking} along with (vi) EGSDG \citep{gu2023train}, , which exploit the edges for generalization. Note that MoreStyle also uses adversarial training. Evaluation followed implementation protocols as reported in the respective publications.

\begin{figure*}
    \centering
    \includegraphics[width=1.0\textwidth]{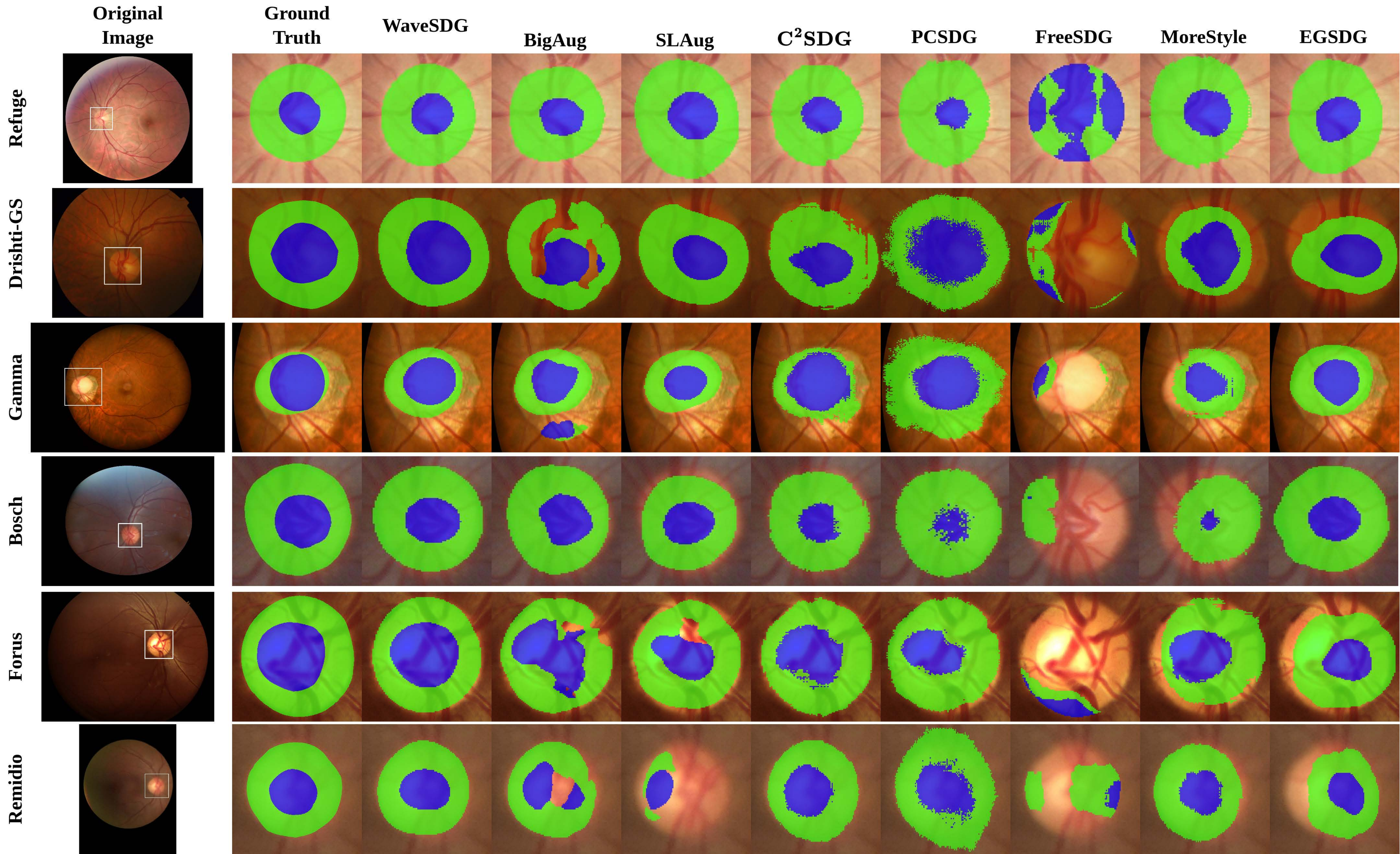}
    \caption{Visualization of predicted segmentation mask by WaveSDG and seven other competing methods, along with the corresponding ground truth. \textcolor{green}{\textbf{Green}} represents the predicted OD region, while \textcolor{blue}{\textbf{blue}} shows predicted OC region. Zoomed view of the cropped region (in white) in the original image, containing the entire OD and OC region, is shown in the remaining columns. Note that for SLAug, false positive regions observed outside the zoomed in region in the Bosch, Forus and Remidio datasets are excluded during visualization.}
    \label{fig:visualres}
\end{figure*}

Table~\ref{tab:quantitativeRes} presents the quantitative results in segmentation of OD and OC, and Fig.~\ref{fig:visualres} shows the corresponding qualitative outputs. Quantitative results show that WaveSDG consistently improved $DSC_{OC}$, while significantly decreasing $HD95$ in almost all target datasets. This demonstrates the ability of WaveSDG to learn structure-dominant and anatomy-aware representations under domain shift. Although baseline models such as BigAug and C$^2$SDG achieved competitive $DSC_{OD}$ with WaveSDG, their performance drop in OC segmentation revealed reduced robustness in capturing subtle and weakly defined region under domain shift. In contrast, WaveSDG achieved high balanced $DSC_{OD}$ and $DSC_{OC}$ in all target data. This indicates that the proposed wavelet-based design suppresses domain-specific variations and learns anatomy-aware semantic representations, while preserving high-frequency edge details required for accurate ROI delineation.

The $HD95$ results provide stronger evidence in support of the strength of the WISER module. WaveSDG achieved the lowest $HD95$ and reduced variance across all domains, indicating stable and precise boundary alignment. The augmentation-based approaches (BigAug, SLAug) produced irregular and fragmented contours (columns 4 and 5 of the figure), and edge-based EGSDG failed to generalize consistently generating under-segmented region in most cases. This demonstrates that neither appearance augmentation nor naive edge modeling is sufficient for generalized segmentation. The improvement observed in both $DSC$ and $HD95$ confirms that WaveSDG captured both region overlap and boundary structures, without introducing over-segmentation artifacts. 

The visual results over sample data in Fig.~\ref{fig:visualres} corroborate the findings over all datasets. The WaveSDG consistently preserves the concentric OD/OC topology while maintaining smooth, coherent boundaries across all domains (column 3 of the figure). In contrast, the competing methods often exhibit failures in boundary localization, structural consistency, and  OC delineation. Augmentation-based method such as BigAug often demonstrate irregular or leaking boundaries and fragment formation, particularly in challenging cases such as Drishti-GS and Gamma. The SLAug, though based on learnable augmentation parameters, fails to capture the topology of the ROI (see column 5 of Bosch, Forus, and Remidio datasets). In the target datasets, the highly competitive $C^2SDG$ displays distorted boundary and under-segmented regions in the Drishti-GS and Forus dataset (see columns 6 and 8 in the figure). MoreStyle fails to identify the OC region (see column 8 for Bosch dataset). Furthermore, Table~\ref{tab:quantitativeRes} and Fig.~\ref{fig:visualres} together emphasize that the WaveSDG outperforms in generalization capability for both standard camera and handheld cameras. The results establish that explicit frequency-aware structural modeling provides a principled and effective solution to DG. 

\subsection{Ablation Studies}
\label{subsec:abal}

 Table~\ref{tab:abal_WISER_DS} demonstrates the role of WISER module and Deep Supervision on the performance of WaveSDG model. It shows that the WISER module consistently improved generalization capability in all target domains. It consistently increased $DSC_{OC}$ and stabilized the $DSC$ score for both classes. This indicates that WISER module improves the representation of fine anatomical structures. Furthermore, a significant reduction in $HD95$ together with high $DSC$, in most target datasets, demonstrates reduced over/under-segmented boundary and robustness to domain shifts. The gains are prominent for OC segmentation. This highlights that the module suppresses domain specific noise and preserves relevant edge information. The addition of Deep Supervision (DS) further amplified these gains, as reflected by the improved $HD95$. DS provided auxiliary gradients to intermediate layers during training. These additional loss signals enforced discriminative feature learning at multiple scales. The combination of WISER and DS yielded best overall results, particularly in challenging target domains such as Drishti-GS and Gamma.

\begin{table}
\centering
\caption{Effect of WISER module and Deep Supervision (DS) on the performance of WaveSDG}
\label{tab:abal_WISER_DS}
\resizebox{0.7\columnwidth}{!}{
\renewcommand{\arraystretch}{1.75}
\begin{tabular}{|lll@{}|c|c|c|c|c|c|}
\hline
\multicolumn{3}{|l|}{Dataset}                                                                                            & \textbf{WISER} & \textbf{DS} & \textbf{$DSC_{OD}$}                & \textbf{$DSC_{OC}$}       & \textbf{$HD95_{OD}$}   & \textbf{$HD95_{OC}$}           \\ \hline
\multicolumn{3}{|l|}{\multirow{3}{*}{\begin{tabular}[c]{@{}l@{}}\textbf{Source} -- \\ Refuge\end{tabular}}}                       & \xmark         & \xmark      & 90.69 ± 4.81                   & 89.31 ± 5.57          & 3.99 ± 2.29            & 3.85 ± 1.92                    \\ \cline{4-9} 
\multicolumn{3}{|l|}{}                                                                                                   & \cmark         & \xmark      & {\ul{90.76 ± 4.59}}             & {\ul{89.68 ± 5.44}}    & {\ul{3.98 ± 2.20}}      & {\ul{3.72 ± 1.81}}              \\ \cline{4-9} 
\multicolumn{3}{|l|}{}                                                                                                   & \cmark         & \cmark      & \textbf{91.21 ± 4.23}          & \textbf{89.76 ± 5.57} & \textbf{3.86 ± 2.20}   & \textbf{3.68 ± 1.78}           \\ \hline
\multicolumn{1}{|l|}{\multirow{15}{*}{{\rotatebox{90}{\textbf{Target}}}}} & \multicolumn{2}{l|}{\multirow{3}{*}{Drishti-GS}}                        & \xmark         & \xmark      & 69.10 ± 20.24                   & 63.07 ± 27.49          & {\ul{15.33 ± 9.33}}     & {\ul{16.95 ± 10.59}}            \\ \cline{4-9} 
\multicolumn{1}{|l|}{}                         & \multicolumn{2}{l|}{}                                                   & \cmark         & \xmark      & {\ul{72.86 ± 13.43}}             & {\ul{64.86 ± 24.01}}    & 18.07 ± 23.39          & 20.32 ± 15.58                  \\ \cline{4-9} 
\multicolumn{1}{|l|}{}                         & \multicolumn{2}{l|}{}                                                   & \cmark         & \cmark      & \textbf{80.57 ± 9.30}          & \textbf{82.06 ± 11.59} & \textbf{9.66 ± 3.97}   & \textit{\textbf{11.44 ± 5.83}} \\ \cline{2-9} 
\multicolumn{1}{|l|}{}                         & \multicolumn{2}{l|}{\multirow{3}{*}{Gamma}}                             & \xmark         & \xmark      & 57.66 ± 6.97                   & 49.05 ± 25.86          & 128.05 ± 104.92        & 122.80 ± 89.34                 \\ \cline{4-9} 
\multicolumn{1}{|l|}{}                         & \multicolumn{2}{l|}{}                                                   & \cmark         & \xmark      & {\ul{70.34 ± 21.04}}             & {\ul{64.34 ± 14.78}}    & {\ul{35.42 ± 30.92}}    & {\ul{41.83 ± 36.19}}            \\ \cline{4-9} 
\multicolumn{1}{|l|}{}                         & \multicolumn{2}{l|}{}                                                   & \cmark         & \cmark      & \textbf{81.49 ± 10.59}          & \textbf{79.52 ± 11.02} & \textbf{10.35 ± 9.33} & \textbf{7.42 ± 7.01}         \\ \cline{2-9} 
\multicolumn{1}{|l|}{}                         & \multicolumn{1}{l|}{\multirow{9}{*}{\rotatebox{90}{\textbf{Chákṣu}}}} & \multirow{3}{*}{Bosch}   & \xmark         & \xmark      & 82.95 ± 5.41                   & 65.99 ± 20.23          & 8.88 ± 3.57            & 9.32 ± 3.82                    \\ \cline{4-9} 
\multicolumn{1}{|l|}{}                         & \multicolumn{1}{l|}{}                        &                          & \cmark         & \xmark      & {\ul{88.62 ± 2.97}}             & {\ul{82.48 ± 9.98}}    & {\ul{4.54 ± 1.68}}      & {\ul{5.17 ± 1.88}}              \\ \cline{4-9} 
\multicolumn{1}{|l|}{}                         & \multicolumn{1}{l|}{}                        &                          & \cmark         & \cmark      & \textbf{90.07 ± 2.95}          & \textbf{84.68 ± 7.52} & \textbf{3.98 ± 1.40}   & \textbf{4.96 ± 1.70}           \\ \cline{3-9} 
\multicolumn{1}{|l|}{}                         & \multicolumn{1}{l|}{}                        & \multirow{3}{*}{Forus}   & \xmark         & \xmark      & 86.30 ± 7.05                   & 77.53 ± 18.50          & 8.22 ± 3.76            & 8.80 ± 4.55                    \\ \cline{4-9} 
\multicolumn{1}{|l|}{}                         & \multicolumn{1}{l|}{}                        &                          & \cmark         & \xmark      & {\ul{90.18 ± 3.27}}             & {\ul{84.41 ± 7.67}}    & {\ul{5.89 ± 2.76}}      & {\ul{6.52 ± 3.26}}              \\ \cline{4-9} 
\multicolumn{1}{|l|}{}                         & \multicolumn{1}{l|}{}                        &                          & \cmark         & \cmark      & \textbf{90.63 ± 2.49}          & \textbf{87.99 ± 5.89} & \textbf{5.14 ± 1.42}   & \textbf{5.41 ± 1.96}           \\ \cline{3-9} 
\multicolumn{1}{|l|}{}                         & \multicolumn{1}{l|}{}                        & \multirow{3}{*}{Remidio} & \xmark         & \xmark      & 74.91 ± 23.77                   & {\ul{76.08 ± 19.93}}    & 11.26 ± 9.17           & 15.57 ± 11.44                  \\ \cline{4-9} 
\multicolumn{1}{|l|}{}                         & \multicolumn{1}{l|}{}                        &                          & \cmark         & \xmark      & \textit{\textbf{85.31 ± 9.27}} & 75.70 ± 18.34          & {\ul{9.44 ± 16.03}}     & {\ul{8.40 ± 4.93}}              \\ \cline{4-9} 
\multicolumn{1}{|l|}{}                         & \multicolumn{1}{l|}{}                        &                          & \cmark         & \cmark      & {\ul{85.29 ± 12.72}}             & \textbf{82.97 ± 12.26} & \textbf{6.35 ± 4.69}   & \textbf{6.74 ± 4.05}           \\ \hline
\end{tabular}}
\end{table}

Table~\ref{tab:distribution_dist} presents the efficacy of the WISER module in reducing the domain gap between source and target datasets. Substantial decrease in MMD and Fréchet Distance indicate that WISER effectively aligned mean embeddings and improved statistical alignment between the source and target datasets. Consistent improvement, across all metrics, signifies improved generalization capability in unseen datasets.

\begin{table}
\centering
\caption{Distances between source and target distributions, with and without WISER}
\label{tab:distribution_dist}
\resizebox{0.4\columnwidth}{!}{%
\begin{tabular}{|l|c|c|}
\hline
\textbf{Metric}  & \textbf{w/o WISER} & \textbf{w WISER} \\ \hline
MMD              & 0.7953                      & 0.6415                    \\ \hline
JSD              & 0.2414                      & 0.2288                    \\ \hline
Fréchet Distance & 686.74                      & 530.01                    \\ \hline
\end{tabular}%
}
\end{table}

\begin{figure}
    \centering
    \includegraphics[width=0.6\columnwidth]{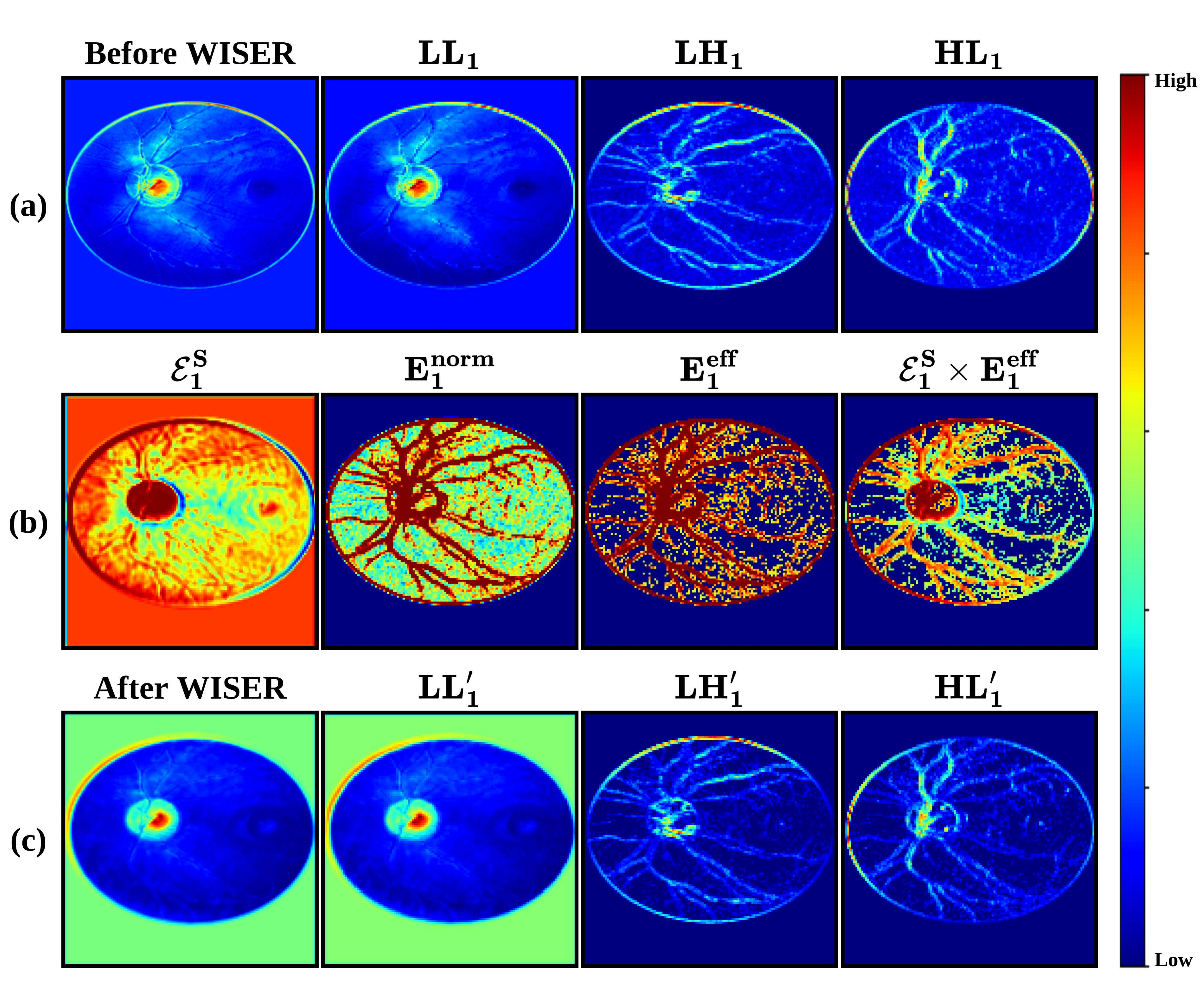}
    \caption{Visualization of the effect of various components in WISER. (a) Feature channel and its wavelet-decomposed sub-bands, prior to the WISER module. (b) Intermediate computation of the scaling factor $\mathcal{E}_1^S$, normalized edge energy $\mathcal{E}_1^{norm}$, boosting factor $\mathcal{E}_1^{eff}$, and their element-wise product. (c) Corresponding WISER-transformed feature channel and filtered sub-bands.}
    \label{fig:abal_wisercomponents}
\end{figure}

Figure~\ref{fig:abal_wisercomponents} visualizes the effect of the Edge Selector (ES) and Edge Booster (EB), within the WISER module [eqns. (\ref{eq:enorm})-(\ref{eq:scalefactor})], on a single wavelet decomposed skip feature channel. The $E_1^{norm}$ captures a broad set of edge-like features inside EB [seen as red, yellow, and light-green region in  row (b) column 2]. Controlled by $E_1^{eff}$, the EB suppresses low-confidence responses (sky blue and green regions in $E_1^{norm}$ are filtered out in $E_1^{eff}$) while retaining the sparse high-energy structures [row (b) column 3]. The $\mathcal{E}_1^S \times E_{eff}$ factor localizes these responses around the salient anatomy. Column 4 of row (b) of the figure shows that the selector controlled by $\mathcal{E}_1^S$ amplifies relevant edges with high-confidence (edges in the ROI have the highest activation, while background edges are suppressed). Post WISER module [row (c)], the $LL_1'$ component becomes smoother and more anatomy-oriented. This indicates that $\varphi_1^c$ learned to capture the relevant anatomy. The $LH_1'$ and $HL_1'$ components retained cleaner and more selective boundary cues. The filtered bands, in the third row of the figure, collectively demonstrate lower background contamination with stronger structural emphasis around the OD/OC regions. These observations indicate that the EB adaptively controls edge enhancement while ES monitors the importance of the edges. The booster and selector collectively convert raw high-frequency responses into anatomy-aware boundary cues. Besides, comparing column 2 of rows (a) and (c) in the figure shows that the transformation on LL disentangles anatomy from its style components. 

\begin{table}
\caption{Resource requirement in WISER}
\label{tab:computation_effi}
\centering
\resizebox{0.45\columnwidth}{!}{%
\begin{tabular}{|l|l|l|l|}
\hline
\textbf{Model} & \textbf{\begin{tabular}[l]{@{}c@{}}Peak Memory\\ (MB)\end{tabular}} & \textbf{\begin{tabular}[l]{@{}c@{}}MACs\\ (G)\end{tabular}} & \textbf{\begin{tabular}[c]{@{}c@{}}\# Params \\ (M)\end{tabular}} \\ \hline
Baseline       & 438.39                                                              & 32.4                                                        & 18.75                                                           \\ \hline
WaveSDG      & 460.40                                                              & 36.7                                                        & 21.99                                                           \\ \hline
               & ($\uparrow$) $5.02\%$                                                            & ($\uparrow$) $13.27\%$                                                   & ($\uparrow$) 3.24                                                              \\ \hline
\end{tabular}%
}
\end{table}

\subsection{Computational load}
\label{subsec:computation_effi}
Table~\ref{tab:computation_effi} shows that the introduction of WISER in the baseline segmentation backbone incurred a negligible $5.02\%$ increase in peak memory, a modest $13.27\%$ computational overhead in terms of Multiply-Accumulate Operations (MAC), and only an additional 3.24~M parameters. This ensures that the WISER module is compatible in resource-constrained settings. WaveSDG maintains a low inference latency of 14.8~ms. Results demonstrate that the WISER module is efficient in terms of resource usage, with superior DG. 

\begin{figure}
    \centering
    \includegraphics[width=0.6\columnwidth]{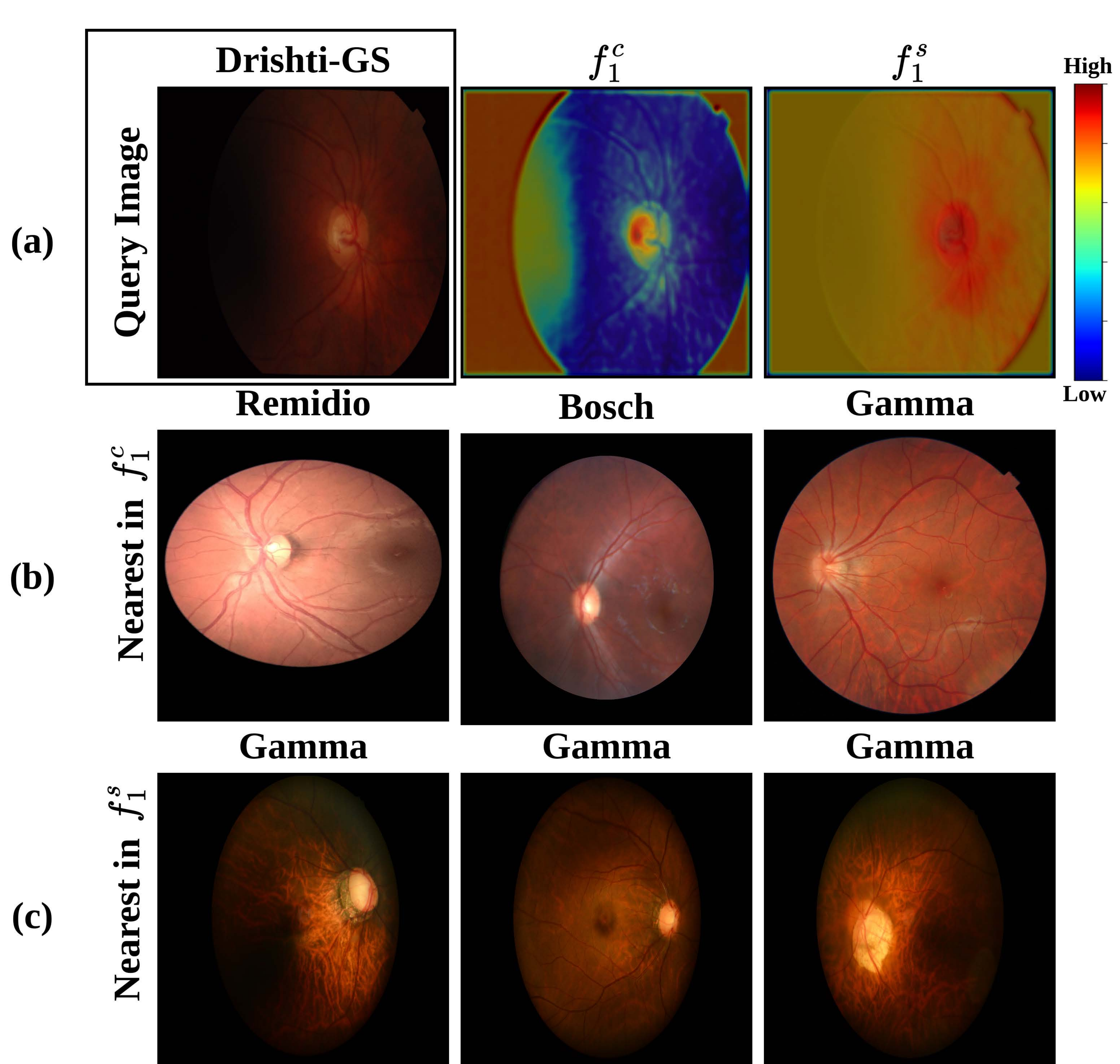}
    \caption{Visualization of LL sub-band decomposition, with row (a) showing column-wise, (i) query image, with its decoupled (ii) anatomy and (iii) style components.  Row (b) displays the respective images, nearest to the query image, in the content embedding space. Row (c) refers to the corresponding images, nearest to the query image, in the style embedding space.}
    \label{fig:LL_decom}
\end{figure}

\subsection{Anatomy-style decoupling}
\label{subsec:visres_anatomy_style}

Figure~\ref{fig:LL_decom} presents an analysis on the filtering of LL sub-band in $\ell = 1$ on a representative query image. The feature responses via $\varphi_1^c$ and $\varphi_1^s$ [of eqn. (\ref{eq:style_content})] are depicted in the row (a), while the row (b) and (c) show nearest neighbors in the content and style embedding spaces, respectively. It is observed that the branch $f_1^c$ produces localized activations aligned with anatomical structures, whereas $f_1^s$ captures appearance-driven responses. The style-space retrieval [row (c)] groups samples with similar low-frequency appearance cues, such as low contrast and vignetting. In contrast, the content space retrieval [row (b)] preserves anatomical similarity in OD location and vessel structure. This confirms the intended decoupling behavior of our WISER module, in terms of $f_1^c$ and $f_1^s$.

\section{Conclusion}
\label{sec:conclusion}
This research presents WaveSDG, a wavelet-guided segmentation framework, that addresses single-source domain generalization in fundus imaging through anatomy-appearance decoupling. The proposed Wavelet-based Invariant Structure Extraction and Refinement (WISER) module leverages the semantic roles of wavelet sub-bands to isolate anatomical structure, refine boundary-defining edges, and suppress noise-driven responses. Incorporation of WISER module in the segmentation backbone, to form WaveSDG, prevents the propagation of domain-specific appearance cues. The results consistently demonstrate superior performance across multiple unseen target datasets, with best balanced Dice scores and lowest 95th percentile Hausdorff distance, as compared to seven SOTA architectures. This study establishes that decoupling anatomy by exploiting the semantic characteristics of the wavelet sub-band provides a promising direction for generalizable segmentation of fundus images. 

The modular design of WISER makes it compatible with most CNN-based encoder-decoder architectures, transferring filtered intermediate features between encoder and decoder layers and improving the generalization capability. Future research can also investigate this strategy under cross-sequence or cross-modality setting.

\section*{Acknowledgment}

Abhirup Banerjee is supported by the Royal Society University Research Fellowship (Grant No. URF/R1/221314).

\bibliographystyle{unsrt}  
\bibliography{references}

\end{document}